\documentclass[conference]{IEEEtran}
\IEEEoverridecommandlockouts
\usepackage{cite}
\usepackage{amsmath,amssymb,amsfonts}
\usepackage{subcaption}
\usepackage{algorithmic}
\usepackage{graphicx}
\usepackage{textcomp}
\usepackage{xcolor}
\def\BibTeX{{\rm B\kern-.05em{\sc i\kern-.025em b}\kern-.08em
    T\kern-.1667em\lower.7ex\hbox{E}\kern-.125emX}}
\begin{document}

\title{Efficient Microscopic Image Instance Segmentation\\ for Food Crystal Quality Control \textsuperscript{*} 
\thanks{\textsuperscript{*} THIS MATERIAL IS BASED UPON WORK SUPPORTED BY THE NATIONAL SCIENCE FOUNDATION UNDER GRANT NO. 2134667.}
}

\DeclareRobustCommand{\IEEEauthorrefmark}[1]{\smash{\textsuperscript{\footnotesize #1}}}

\author{\IEEEauthorblockN{Xiaoyu Ji\IEEEauthorrefmark{1}, Jan P Allebach\IEEEauthorrefmark{1}, Ali Shakouri\IEEEauthorrefmark{1}, and Fengqing Zhu\IEEEauthorrefmark{1}}

\IEEEauthorblockA{\IEEEauthorrefmark{1}Elmore School of Electrical and Computer Engineering, Purdue University, West Lafayette, IN 47907, USA} 
\IEEEauthorblockA{\{ji146, allebach, shakouri, zhu0\}@purdue.edu}}

\maketitle

\begin{abstract}
This paper is directed towards the food crystal quality control area for manufacturing, focusing on efficiently predicting food crystal counts and size distributions. Previously, manufacturers used the manual counting method on microscopic images of food liquid products, which requires substantial human effort and suffers from inconsistency issues. Food crystal segmentation is a challenging problem due to the diverse shapes of crystals and their surrounding hard mimics. To address this challenge, we propose an efficient instance segmentation method based on object detection. Experimental results show that the predicted crystal counting accuracy of our method is comparable with existing segmentation methods, while being five times faster. Based on our experiments, we also define objective criteria for separating hard mimics and food crystals, which could benefit manual annotation tasks on similar dataset.

\end{abstract}

\section{Introduction}

Microscopic images play an important role in manufacturing crystallization process assessment, including food and electronic products \cite{CHAO2010,ZHU2019,KIANI2011}. The product we analyzed in this paper is a food product in liquid form; and microscopic images are captured at each step of the processing pipeline for quality control. The quality of samples captured by a microscope is related to the size and density of food crystals. Previously, manufacturers count crystals manually and estimate the crystal density with the naked eye. Food crystals in the microscopic images are similar to some noise or blurry objects, which is difficult for people to handle consistently during manual counting. To save time and keep the analysis process consistent, an automatic and accurate crystal segmentation program is necessary for the microscopic image analysis.

\begin{figure}[t]
  \centering 
   {\includegraphics[scale=0.55]{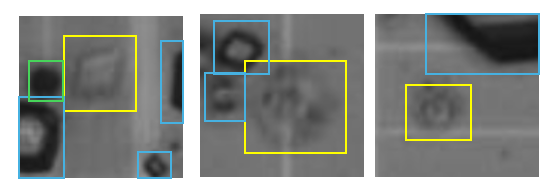}}
    \caption{Hard mimics examples. The objects in yellow boxes are hard mimics. Crystals and air bubbles are in blue and green boxes, respectively.}
    \label{fig:11}
\end{figure}

The image processing program proposed in \cite{Qiyue2023} is an earlier approach to deal with images captured by a low-resolution microscope. As the device has been updated, the image resolution is increased and the parameters of the image processing program have not been adapted to the new images. In this paper, we explore deep learning methods to improve the prediction accuracy. However, pre-trained instance segmentation models using public datasets such as ImageNet \cite{Deng2009} dataset and DSB2018 dataset \cite{Caicedo2019} do not work well on our data. Our dataset is annotated into three classes: air bubbles, hard mimics, and crystals, where air bubbles can be easily separated. Hard mimics represent the out-of-focus objects, noise, and airborne dust in the images. The hard mimics are of various shapes and similar size as the real crystals, as shown in Fig. \ref{fig:11}.

For consistent manual annotation, setting the criteria to distinguish hard mimics and food crystals is very important. However, the experiences of long-time trained operators are not written into standards. In this paper, we set fixed criteria about the definition of hard mimics to help us understand the difference between crystals and hard mimics, and maintain the consistency of manual ground truth labeling. All the criteria are learned from the experimental descriptions of the manufacturers and these criteria can also benefit similar microscopic dataset annotation.

The data we used is named the seed count food crystal dataset. The ``Seed Count" stage is the first stage of all the processing steps, which is the most important one for manufacturers to decide whether an early stop is needed if the product is of bad quality. The most significant assessment item is the crystal counts per square millimeter, while the size of the crystals and the area covered by crystals are also essential for quality control. In this paper, we use deep learning based method to improve the prediction accuracy of these metrics.

Deep learning has been widely used in image instance segmentation in recent years \cite{gu2022,ZHANG2023,he2017}. Image instance segmentation separates objects of the same class and predicts a pixel-wise mask for each object. The existing image segmentation methods such as Mask R-CNN \cite{he2017}, Mask-scoring R-CNN \cite{huang2019mask}, and Stardist \cite{schmidt2018cell} outperform the prior image processing method we used \cite{Qiyue2023}. However, the inference process of the segmentation methods is slow. To speed up the prediction and improve the counting accuracy, we explored a detection-based method and designed a post-processing program to segment crystals pixel-wisely. The prediction accuracy of our proposed method improves somewhat over the existing segmentation methods; and it is much more efficient in terms of computation time.

\section{Related Work}

In this section, we review existing works related to image instance segmentation and food crystal image analysis.

\subsection{Instance segmentation}

Image segmentation maps one image into the pixel-wise mask of different object classes inside. Instance segmentation, one sub-task of image segmentation, predicts objects from the same class into individual masks instead of one mask. It has been used in both 2D and 3D image analysis areas such as medical image segmentation \cite{zhou2019cia,qian2022transformer}, virtual reality \cite{zhang2020virtual,han2020occuseg}, autonomous driving \cite{zhou2020joint,mohamed2021monocular} and so on. Due to the remarkable achievement of deep neural networks, in this section, we focus on the instance segmentation methods using deep learning technology.  

Instance segmentation can be separated into ``detect and segment" models and directly segment models. One-step segmentation models predict pixel-wise instance masks directly based on the latent representatives learned by the backbone network. Stardist \cite{schmidt2018cell} is a widely used method for nuclei instance segmentation. The model combines a U-net architecture with fully connected layers to predict the polygon shape object probability and distance from the edge to the object center. 

The ``detect and segment" models predict the bounding box region of objects first with object detection model, and then pixel-wisely segment the output masks. For example, Mask R-CNN \cite{he2017} predicts the object bounding box with the region proposal network, and then highlights the box region of the predicted feature maps for classification and pixel-wise segmentation. Further, Mask-scoring R-CNN \cite{huang2019mask} is proposed to improve the prediction details, which calibrates the inaccurate mask confidence score by estimating the quality of each predicted instance mask. In our paper, we use a different method to realize the ``detect and segment" idea. The object detection and pixel-wise segmentation are separated. We combine the deep learning object detection method, You Only Look Once (Yolo) framework \cite{yolov8}, with image processing segmentation techniques to improve the prediction accuracy.

\subsection{Food crystal image analysis}

Food crystal samples are mostly found in liquid and solid food products, such as milk, cheese, and sugar \cite{schumacher2015automated,WU2023111435,impoco2011segmentation}. The automatic image analysis results not only provide valuable evaluation metrics but also benefit crystallization process prediction \cite{velazquez2010characterization,wu2023sensor}. The crystals in the microscopic images of each product have different shapes and size distributions. It is hard to directly apply a model developed for one food crystal image dataset to a different food crystal image dataset. Existing segmentation works related to food crystal image analysis can be split into image processing methods and deep learning methods. For example, ASSugarNet has been proposed \cite{WU2023111435} to improve overlapping crystal prediction with a pyramid deep neural network. Other studies such as \cite{schumacher2015automated, Qiyue2023} implement automatic image processing pipelines to realize image segmentation. In our study, we focus on predicted counting accuracy instead of overlapped crystal sizes, and prior image processing works achieve poor results compared to existing deep learning segmentation methods. 

\section{Method}
In this section, we describe our model architecture including the object detection model and the post processing module.

\begin{figure*}[t]
  \centering
  \includegraphics[scale=0.5]{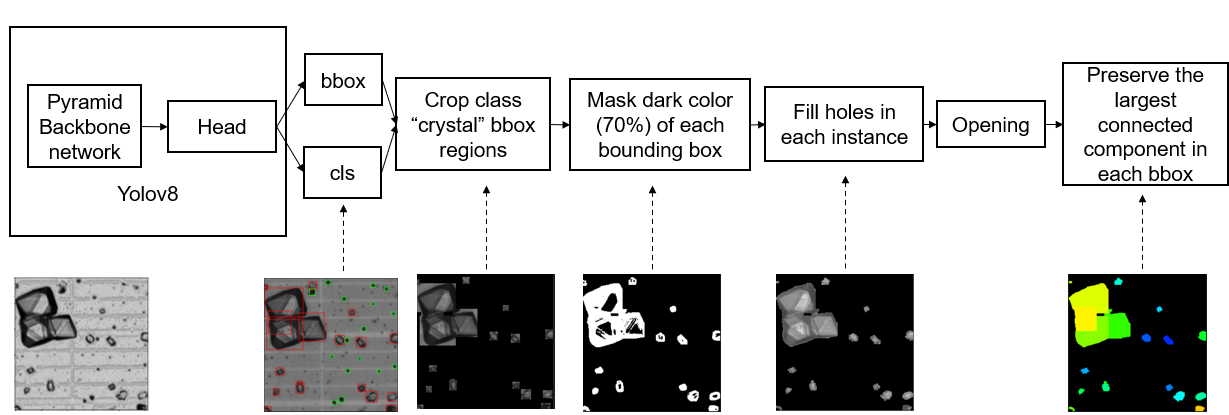}
  \caption{Overview of our proposed method. ``Bbox" represents bounding box, ``cls" represents class.  The images from left to right are an example input image and the results of each step. }
  \label{fig:1}
\end{figure*}

\subsection{Object detection model}

Yolov8 is used here as the object detection model \cite{yolov8}. As shown in Fig. \ref{fig:1}, the architecture includes a five-layer pyramid backbone network and a Yolov8 head that extracts the output features of the top three layers in the backbone network. The predicted bounding box (bbox) belongs to one of the three classes (cls): crystal, hard mimic, and air bubble.

\subsection{Post processing}

The post-processing pipeline shown in Fig. \ref{fig:1} includes five parts. We process the crystal class predictions and crop out the bounding box areas. To separate the background color and the crystal edges, binarization is applied by assigning the 70\% darkest pixels within each bbox region as 1 and others as 0. This is an experimental threshold we tested on our dataset. Next, we perform a set of morphological operations. We use an infilling operation to fill the holes inside the edges of crystals. After this step, the extracted edges for each crystal are thicker than the actual edges, so we use an opening operation to remove the extra edge pixels. Finally, since each bbox can include the boundary or corner of other crystals, we only preserve the largest connected component in each bbox to remove the residual areas.

\section{Dataset and Manual Annotation Criterion}

\subsection{Seed count food crystal dataset}
The dataset we used was captured from a high-resolution microscope device during the seed count manufacturing stage. The liquid product is dropped on one slide of glass and covered with another. During this operation, inevitable air pockets and dust particles form between the glass and the liquid, which generates air bubbles and some of the hard mimics.  

There are two datasets we used in this study. \textit{Dataset 1} is a pixel-wise labeled dataset comprising 171 training images and 96 test images, each sized 256$\times$256. All the annotations were completed by one operator, encompassing all three classes: crystal, hard mimic, and air bubble. \textit{Dataset 2} only includes the total ground truth crystal counts of each image instead of pixel-wise labels, consisting of 35 images with size 1536$\times$2048, which were generated by a more experienced operator different from the operator who generated the \textit{Dataset 1} annotations. Because there is potential individual bias in the annotations of the first dataset, we use the second dataset to validate our results.

\subsection{Manual annotation criteria}

\begin{figure}[t]
  \centering 
   {\includegraphics[scale=0.6]{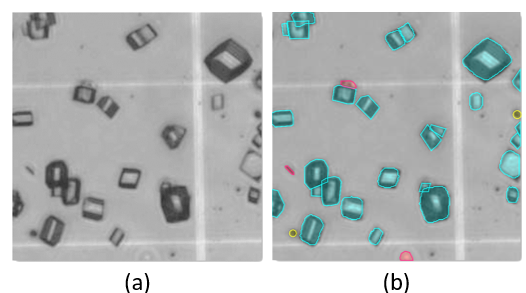}}
    \caption{Example food crystal images captured from the microscope. (a) original image (lightness adjusted), (b) manual annotation result. Crystals are marked with blue, air bubbles are marked with yellow, while hard mimics are marked with red.}
    \label{fig:2}
\end{figure}

As shown in Fig. \ref{fig:2}, the instances marked with blue are crystals while those marked with red are hard mimics. Hard mimics have blurry edges or non-polygon shapes, but the difference between hard mimics and crystals needs additional differential criteria. The criteria we set are listed below:

\begin{itemize}
\item Hard mimics have a non-polygon shape with at least one edge invisible or blurry
\item An object with all faint edges is a hard mimic
\item Small items with no openings in them are hard mimics
\item Items on the boundary with no openings and very small items are hard mimics
\end{itemize}

All the criteria are derived from the descriptions of experienced operators. The ``non-polygon shape" in the first bullet and ``small items" in the third bullet of the criteria are used to separate crystals from out-of-focus items and noise.

\section{Experiments}

\begin{figure*}[!ht]
  \centering
   \subcaptionbox
      {Input image\label{result:a}}{\includegraphics[scale=0.83]{ 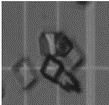}}
    \subcaptionbox
      {Ground truth\label{result:b}}{\includegraphics[scale=0.83]{ 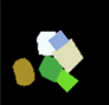}}%
       \hspace{1ex}
    \subcaptionbox
      {Mask RCNN \cite{he2017}\label{result:c}}{\includegraphics[scale=0.83]{ 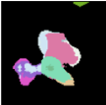}}
     \hspace{0.1ex}
    \subcaptionbox
      {Mask-scoring RCNN \cite{huang2019mask}\label{result:d}}{\includegraphics[scale=0.83]{ 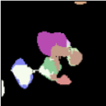}}%
      \hspace{0.5ex}
    \subcaptionbox
      {Stardist \cite{schmidt2018cell}\label{result:e}}{\includegraphics[scale=0.85]{ 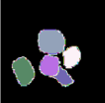} }
    \subcaptionbox
      {Yolov8 segmentaton \cite{yolov8}\label{result:f}}{\includegraphics[scale=0.25]{ 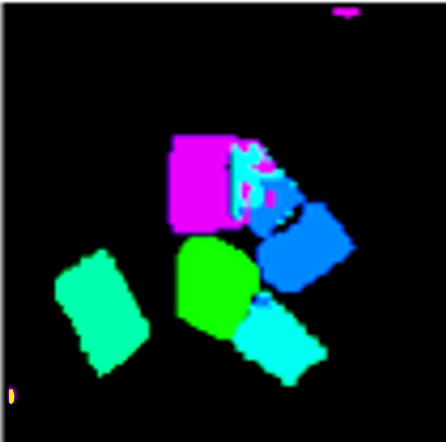} }%
      \hspace{0.5ex}
    \subcaptionbox
      {Ours\label{result:g}}{\includegraphics[scale=0.83]{ 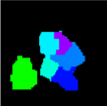}}%

\subcaptionbox
      {Input image\label{result:h}}{\includegraphics[scale=0.71]{ 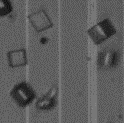}}
    \subcaptionbox
      {Ground truth\label{result:i}}{\includegraphics[scale=0.71]{ 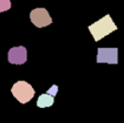}}%
       \hspace{1ex}
    \subcaptionbox
      {Mask RCNN \cite{he2017}\label{result:j}}{\includegraphics[scale=0.71]{ 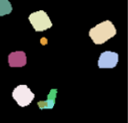}}
     \hspace{0.1ex}
    \subcaptionbox
      {Mask-scoring RCNN \cite{huang2019mask}\label{result:k}}{\includegraphics[scale=0.71]{ 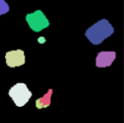} }%
      \hspace{0.5ex}
    \subcaptionbox
      {Stardist \cite{schmidt2018cell}\label{result:l}}{\includegraphics[scale=0.71]{ 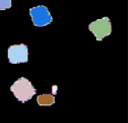} }
    \subcaptionbox
      {Yolov8 segmentaton \cite{yolov8}\label{result:m}}{\includegraphics[scale=0.71]{ 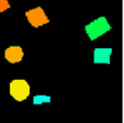} }%
      \hspace{0.5ex}
    \subcaptionbox
      {Ours\label{result:n}}{\includegraphics[scale=0.71]{ 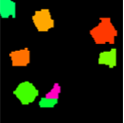}}%
  \caption{Visualization comparisons of different segmentation methods on food crystal dataset. From left to right are the input images, the ground truth label images, results of Mask RCNN \cite{he2017}, Mask-scoring RCNN \cite{huang2019mask}, Stardist \cite{schmidt2018cell}, Yolov8 segmentaton \cite{yolov8} and our method.}
  \label{fig:3}
\end{figure*}

\subsection{Experimental Setting}

We compared our method with several existing image segmentation methods. All the models are trained using the training images from \textit{Dataset 1} (171 images with size 256$\times$256), given ground truth labels of the three classes. The training and testing processes are performed on an NVIDIA GeForce GTX 1080 Ti GPU.

We trained the detection model with 1000 epochs using the Adam optimizer \cite{Kingma2014AdamAM} with $\beta1 = 0.937$, $\beta2 = 0.999$ and $w=5\times10^{-4}$. The learning rate was initialized as 0.01. The box loss gain was set at 7.5, and the class loss gain at 0.5. The scale of the model is Yolov8 large.

\subsection{Quantitative Results}
\label{sec:exp}

We test the models on the 96 test images in \textit{Dataset 1}. As shown in Table. \ref{tab:table_1}, there are five evaluation metrics used. The three metrics on the left are the measurements that manufacturers pay attention to. CntAcc represents counting accuracy, CovErr represents the crystal-covered area error, and SizeErr represents the mean crystal size error in the units of micronmeters. The detailed equations are listed in (\ref{eq:1}). The last two metrics are segmentation accuracy metrics; and both used an intersection over union (IoU) confidence threshold of 50\%. mAP50 represents mean average precision, recall50 is the recall metric. 

\begin{equation}
\centering
\begin{aligned}
\label{eq:1}
&CntAcc=1-\frac{1}{n}\sum_{i=1}^{n}||\Tilde{cnt}-cnt||/cnt\\
&CovErr=\frac{1}{n}\sum_{i=1}^{n}||\Tilde{cov}-cov||/cov\\
&SizeErr=\frac{1}{n}\sum_{i=1}^{n}||\Tilde{msize}-msize||\\
\end{aligned}
\end{equation}

In Equation (\ref{eq:1}), $n$ is the number of test images. $\Tilde{cnt}$ and $cnt$ are the predicted and ground truth crystal counts, respectively. $\Tilde{cov}$ and $cov$ are the predicted and ground truth crystal-covered area of each image. $\Tilde{msize}$ and $msize$ are the predicted and ground truth average crystal size of each image. 

\begin{table}[htbp]
\caption{Evaluation metrics of each method tested on \textit{Dataset 1}}
\centering  
\label{tab:table_1}
\normalsize
\begin{tabular}{|p{0.2\columnwidth}|p{0.1\columnwidth}|p{0.1\columnwidth}|p{0.1\columnwidth}|p{0.11\columnwidth}|p{0.11\columnwidth}|} 
\hline
Models & CntAcc (\%) $\uparrow$ & CovErr (\%) $\downarrow$ & SizeErr $\downarrow$ & mAP50 (\%) $\uparrow$ & recall50 (\%) $\uparrow$\\
\hline
\hline
Mask RCNN \cite{he2017} & 81.865 & 2.576 & \textbf{1.295} & 74.61 & 82.76\\
\hline
Mask-s RCNN \cite{huang2019mask} & 89.941 & 1.548 & 1.365 & 80.76 & 82.06\\
\hline
Stardist \cite{schmidt2018cell} & 76.849 & 1.962 & 1.551 & \textbf{89.86} & 68.86\\
\hline
Yolov8 segm \cite{yolov8} & \textbf{92.915} & \underline{1.263} & 2.367 & 74.57 & \textbf{91.45}\\
\hline
Ours  & \underline{92.573} & \textbf{1.050} & \underline{1.330} & \underline{83.29} & \underline{87.01}\\
\hline
\end{tabular}
\end{table} 

As shown in Table \ref{tab:table_1}, the bolded numbers are the best value for each metric, and the under-lined numbers are the second-best value. Segmentation models (first four models in Table \ref{tab:table_1}) have different performances due to the difference in model architectures. We focus on analyzing the difference between our method and segmentation models.

For the most important metric CntACC, the Yolov8 segmentation model achieves the best performance while our method is the second-best performing one. The detection module of our method is trained on the bounding box labels, which focuses on crystal location prediction. CntAcc metric represents the accuracy of the detection module. Compared to segmentation models (first four models in Table \ref{tab:table_1}) trained on pixel-wise instance labels, our experiment shows that the detection module requires less information from the ground truth and achieves similar performance on the CntAcc metric.

However, the Yolov8 segmentation model has a worse performance on the CovErr and SizeErr metrics than ours, as shown in Table \ref{tab:table_1}. These metrics are also highly related to the food crystal quality. The CovErr and SizeErr metrics of our method benefit from the post-processing module. Compared to segmentation models that directly predict pixel-wise masks, our two-step architecture efficiently segments pixel-wise information in the second step and achieves comparable results to other methods.

The last two columns in Table \ref{tab:table_1} measure instance-wise segmentation accuracy, mAP50 represents the percentage of predicted crystals that are accurate while recall50 represents the percentage of true crystals that are predicted. There is a trade-off between mAP50 and recall50, and either being low is a strong weakness. Compared to the segmentation models, our method balances the two metrics by separating detection and pixel-wise segmentation. The detection module accurately predicts the location of instances, while post-processing achieves high IoU between instances.

In general, our method has the best or the second-best performance for all the metrics, while other methods have their obvious shortcomings.

To evaluate our method more objectively, we test on the 35 images in \textit{Dataset 2}. Table \ref{tab:table_2} shows the quantitative results of the best three models from previous experiments. The crystal count accuracy of our method is also the highest of all the methods.  

\begin{table}[htbp]
\caption{Evaluation metrics of each method tested on \textit{Dataset 2}}
\centering  
\label{tab:table_2}
\normalsize 
\begin{tabular}{|p{0.20\columnwidth}|p{0.22\columnwidth}|p{0.22\columnwidth}|p{0.10\columnwidth}|p{0.15\columnwidth}|p{0.10\columnwidth}|} 

\hline
Metrics & Stardist \cite{schmidt2018cell} & Yolov8 segm \cite{yolov8} & ours\\
\hline
CntAcc (\%) & 88.88 & 87.42 & \textbf{89.20}\\
\hline
\end{tabular}
\end{table} 

\subsection{Processing Time Comparison}

The time consumption of different methods is evaluated on the GPU that was indicated previously as the platform used for the experiments. We test the best-performing three models in Section \ref{sec:exp}. The existing segmentation methods take 32-37 seconds for one image of size 1536$\times$2048, as shown in Table \ref{tab:table_3}, while our proposed method takes only 5.94 seconds. The object detection model takes 3.67 seconds, and the post-processing pipeline takes 1.82 seconds. The processing time of the proposed method is about five to six times faster than the existing segmentation methods.

\begin{table}[htbp]
\caption{Time consumption of each method test on one image}
\centering  
\label{tab:table_3}
\normalsize
\begin{tabular}{|p{0.15\columnwidth}|p{0.22\columnwidth}|p{0.22\columnwidth}|p{0.10\columnwidth}|p{0.10\columnwidth}|} 
\hline
 & Stardist \cite{schmidt2018cell} & Yolov8 segm \cite{yolov8} & ours\\
\hline
time (s) &  32.04 & 36.50 & 5.94\\
\hline
\end{tabular}
\end{table}

\subsection{Qualitative Comparison}

In this section, we visually compare the segmentation results of different methods. There are five prediction results in Fig. \ref{fig:3}. Fig. \ref{result:a} and Fig. \ref{result:h} are the original images. Fig. \ref{result:b} and Fig. \ref{result:i} are the ground truth labeled images. From the frist row, we compare the number of crystals in the cluster. Fig. \ref{result:c} and Fig. \ref{result:d}, which are the results of Mask RCNN \cite{he2017} and Mask-scoring RCNN \cite{huang2019mask}, include more crystals than the ground truth. The Stardist \cite{schmidt2018cell} result shown in Fig. \ref{result:e} has one crystal missing. The Yolov8 segmentation \cite{yolov8} result shown in Fig. \ref{result:f} includes two small hard mimics on the top and left edges. In the result image of our method (Fig. \ref{result:g}), the number of crystals matches with the ground truth. In the second row, Fig. \ref{result:j} and Fig. \ref{result:k} incorrectly detected a bubble. Fig. \ref{result:l} has one crystal missing on the right side of the image, while Fig. \ref{result:m} has one missing on the bottom. To conclude, our method has a higher counting accuracy than other methods.

\subsection{Limitation and Future Works}

\begin{figure}[t]
  \centering 
   {\includegraphics[scale=0.38]{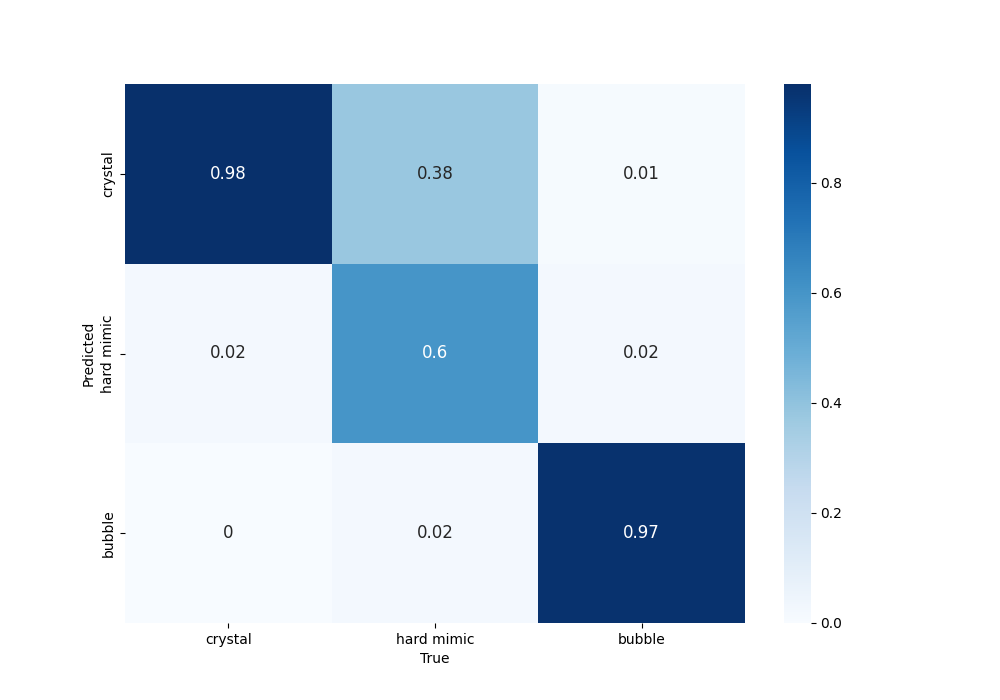}}
    \caption{Normalized confusion matrix for the object detection model.}
    \label{fig:4}
\end{figure}

Our proposed method has the following two limitations. First, the object detection classification accuracy can be further improved. As shown in Fig. \ref{fig:4}, the prediction accuracy of the hard mimic class is much lower than that of the crystal class and the bubble class. The second limitation is that very large crystals fall outside the data distribution learned by the model, which are then predicted as multiple small pieces around the edges. This is an important issue to solve in our future work. We would further work on the images from the later stages of the manufacturing pipeline and address the crystal overlapping issue.

\section{Conclusion and Future Work}

In this paper, we propose an efficient food crystal microscopic image segmentation method. We improve the crystal counting accuracy, which is the most important metric that manufacturers care about, with our object detection based method. The architecture proposed combines existing detection deep neural networks with a post-processing image processing module. The processing time is five to six times faster than the existing segmentation methods. The program proposed benefits the daily life of the manufacturer with whom we are collaborating.

\bibliographystyle{IEEEbib}
\bibliography{reference}

\end{document}